# X-AXIS BASED REGRESSION

February 12, 2020

Dylan Wilson

**Abstract:** In this paper, I will introduce a new form of regression, that can adjust overfitting and underfitting through, "distance-based regression". Overfitting often results in finding false patterns causing inaccurate results, so by having a new approach that minimizes overfitting, more accurate predictions can be derived. Then I will proceed with a test of my regression form and show additional ways to optimize the regression. Finally, I will apply my new technique to a specific data set to demonstrate its practical value.

**CONTENTS**

**Introduction**  In this paper I will introduce a new form of regression, "Overfitting Based Regression" which allows you to tune the level of overfitting or underfitting, with the goal of generalizing standard regression methods.  This new regression technique produces a nonlinear function of the x or right hand side variables using weights on neighboring data points, instead of the traditional approach of applying the best fit line.  The method approximates the dependent value by using a weighted average of the dependent variable based on distances to the point which you are attempting to estimate. It does this for three critical reasons: first because this eliminates the need for arbitrary best fit lines, second with this type of "enhanced" regression can adjust the level overfitting to further optimize your results, and third, this allows "overfitting based regression".  This regression addresses several flaws  inherent  to traditional regression analysis: giving too much freedom to the best fit line, allowing it to find patterns that are not there through overfitting, or the inverse miss existing patterns; therefore, this new model specifically measures these two characteristics and seeks to effectively balance them.  More importantly,  it eliminates the need to employ arbitrary best fit functions, linear, log, polynomial or exponential which are often built upon unnecessary assumptions, resulting in less accurate predictions. Furthermore, this new form of regression eliminates problems that result from heteroscedasticity.

**1. Distance and X-axis Based Regression** In this chapter I will introduce the idea of an "X-Axis Based Regression ", then I will give the most basic example, a regression that minimizes the sum of the distance from all the points.  Then I will set a framework in order to formalize this model in turn allowing this framework to be optimized to create the most effective model possible.

**1.1 X-Axis Based Regression** The idea of  "X-Axis Based Regression" is that instead of trying to produce a "generic" regression that solves for the set of all solutions, it solves for X value on an individual basis. Given a X-value, the model estimates the Y value. First,  the X  value that is being solved is defined by the equation, $Y_\chi = Z_0$, where Z represents the optimal value of Y.  The goal of the regression is to find the value $Y_\chi$ that best estimates an unknown semi random function, $Z_0$. This can be quantified in a cost function by measuring the distance between the predicted and and the actual function output of the function for X

$$Y_x = \{Z\}, \ min(f(X, \sim U([-\infty, \infty])) - Z_0)$$

Where Z is the value of $Y_x$ that is closest to the output of the semi-random function $(f(X, \sim U([-\infty, \infty])))$. This criteria states that given a function distributed across Y values, we want to find the point which reduces the the distance between all the outcomes along the given line X to create an optimal prediction $Z_0$. Given this criteria, we can create a new criteria in respect to the given dataset to solve for $Z_0$. Since the optimal function is unknown, the other y values can be used to estimate the distribution of points along this given axis; however, the past distribution of Y values does not accurately represent the y values along the given X axis. However, the closer the x value the more accurate it represents the theoretical distribution of points across the X-axis. Therefore, you can created a weighted average by measuring its closeness in X-value, w $(X - X_i)$. This model best predicts the theoretical distribution of Y values across this X-axis, because the closer the x value the more representative of the actual distribution of y values, therefore, resulting in a weighted distribution of points representing the most likely y distribution. Given this weighted distribution of y values we can find a point that best predicts our dataset. This approach results in a weighted average by some inverse function of distance: w $(X - X_i)$. This can be formalized as:

$$min(\sum_{i=1}^{n} w(X - X_i)(Z_0 - Y_i)^2)$$

The cost function w is used to solve for the value of Z given monotonically decreasing function for $w(X - X_i)$ creating a weighted average by distance. Z: the variable adjusted to reduce the cost function. $X_i$: the set of the X values for known points on the graph, $Y_i$: the set of the Y values for known points on the graph, and X: the x value of that model attempts to estimate. The cost function can easily be altered to be able to include more independent variables:

$$min(\sum_{i=1}^{n} f((Z_0 - Y_i), (D_1 - D_{i1}), (D_2 - D_{i2}), (D_3 - D_{i3}) ... (Dn - D_i, n))$$

The above function can combine the variables through vectors. This formula above generalizes all possible cost functions. This cost function can be solved through three known methods: estimating through gradient descent, finding the derivative of the function, then finding the minimums and maximums of the function, and through simplifying summations through linearity. The gradient descent provides a general solution but takes an infinite number of calculations to find the *true* or *exact* solution, however, almost any personal computer can solve for a near perfect solution in milla-seconds, therefore it still is an effective and practical alternative to other regressions.

**1.2 Distance Based Regression** Distance based regression is a regression that uses a similar cost function to most regression forms though instead of controlling the bounds of the regression through a best-fit line, the distance based regression limits the set of answers to a single x-axis. The cost function that linear regression uses is the sum of the distances from the line to the known data points, similarly distance based regression reduces the sum of the distances from some point $Y_\chi = \{Z\}$ to the known points. This can be formally stated as;

$$\min(\sum_{i=1}^{n} \sqrt{(X - X_i)^2 + (Z_\varrho - Y_i)^2})$$

The equation above is minimized in respect to $Z_\varrho$, for which $X_i$ is the array of dependent variables and $Y_i$ is the set of dependent variables and X is the instance for which you are trying to solve. Shown in a more intuitive way in the diagram below, $Z_\varrho$ in this diagram is the point along the line 0.6 for which you are trying to solve which minimizes the sum of the distance from the set of all the other points.

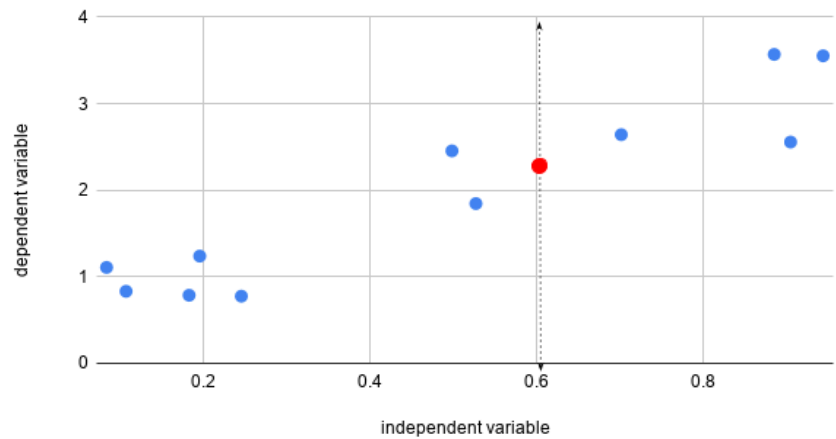

We can start by

$$\sum_{i=1}^{n} \sqrt{(X - X_i)^2 + (Z_\varrho - Y_i)^2}$$

In order to solve this type of problem the derivative is taken to find the 0 derivatives resulting in the relative minimums. Then take the derivative with respect to $Z_\varrho$, the independent variable.

$$\sum_{i=1}^{n} (X - X_i)/\sqrt{(X - X_i)^2 + (Z_\varrho - Y_i)^2} = \sum_{i=1}^{n} \cos\Theta_i = 0$$

The x distance to the optimal point divided by the euclidean distance is equivalent to the cos (theta)  created by the line from the theoretical optimum to itself. However, the sum of the cosines equaling zero is equally hard to solve; however, we can simplify this by relating it back to the known variables and from there finding the slope.

$$\sum_{i=1}^{n} \cos(\tan^{-1}(\frac{Z_\varrho - Y_i}{X - X_i})) =$$

Using the Pythagorean and theorem it can be can derive

$$\sum_{i=1}^{n} \cos(\tan^{-1}(\frac{Z_\varrho - Y_i}{X - X_i})) = \sum_{i=1}^{n} \frac{1}{\sqrt{(\frac{Z_\varrho - Y_i}{X - X_i;n})^2 + 1}} = 0$$

But even so, there is no clear analytical solution.

The minimization problem involves minimizing $\sum_{i=1}^{n} di$ over Z. The first order condition for a

minimum requires $\sum_{i=1}^{n} di^{\wedge}(-1)$ (Z-Yi), which gives

$$Z = \frac{\sum_{i=1}^{n} Yi/di_i}{\sum_{i=1}^{n} 1/di}.$$

The right hand side of this expression is a function of Z through $di$, but one can take a candidate Z and iterate until a fixed point solution is found.

However, using distance based regression is very easy to solve for using Euler's method and gives a decent estimation of a semi-random function. This can be seen in the following graph where a curve was given with a probability distribution applied and resulted in the following graph where the blue points are the estimates and the orange are the given data. This result is not particularly compelling, however, this estimation is using very few points.

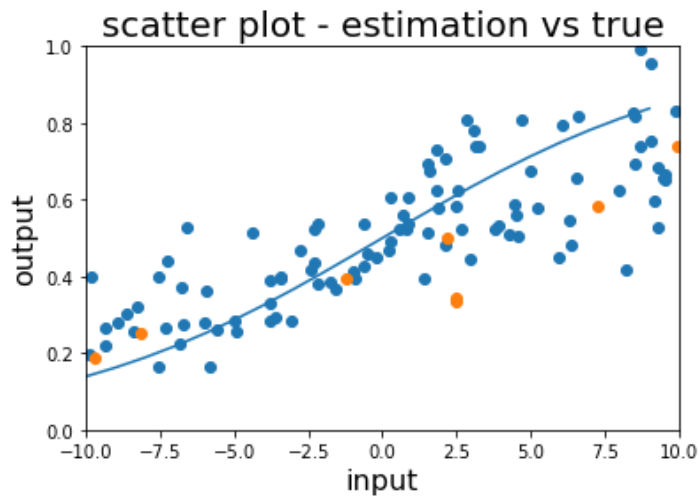

Given this test with more points:

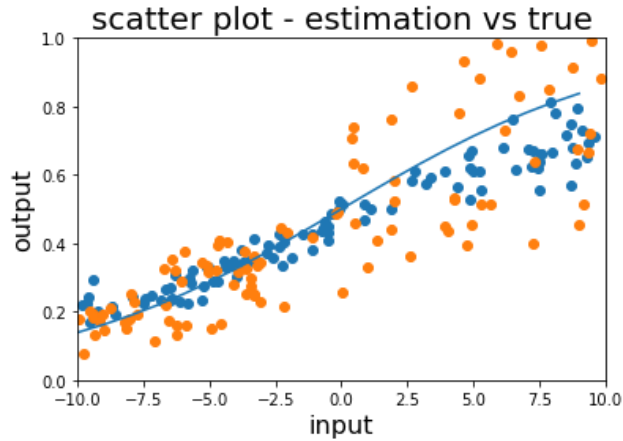

This performance is exceptional considering that the only weights used were those arising naturally from the euclidean distance metric, and that nearby points are not weighted more heavily than more distant points., and there was a large amount of noise included in the artificial test. Furthermore, it only degenerated on the edges where there is significantly less data. However, the blue points representing the estimates for the graph are highly variant. The variance can be controlled through weights.

**2. Weighted Regression** In this chapter, I will continue on the idea of X-axis based regression by introducing different cost functions and their respective solutions. Then I will proceed to find the most effective cost function given a data set through quantifying the variance.

**2.1 Division Based Weights** The division based cost function minimizes the y distance from other points weighted by the distance of the x difference for those points. It does this by applying a cost function that divides some function of the y distance by some function of the x distance resulting in closer x values having more significance in the cost function. Using a function that decreases importance relative x difference is logical, as it creates an estimate that is more similar to the surrounding point along the graph of the function. The cost function division based weights can be formalized as a sum as;

$$min(\sum_{i=1}^{n} f_2(Z_\varrho - Y_i)/f((D_1 - D_{i1}), (D_2 - D_{i2}), (D_3 - D_{i3})...(Dn - D_i, n))$$

The most simple form of this can be written as;

$$min(\sum_{i=1}^{n} (Z_\varrho - Y_i)^2 / (x - x_i)^2)$$

This is a weighted regression where the weights are the inverse of the squared distance. Taking the product of the denominators and multiply the equation by this as the minimum is still the same

$$min((\sum_{i=1}^{n} (Z_\varrho - Y_i)^2 / (x - x_i)^2)=$$

$$\sum_{i=1}^{n} ((Z_\varrho - Y_i)^2 (1/(x - x_i)^2))$$

$(\prod_{i=1}^{n} [(x - x_i)^2])/(x - x_i)^2$ —*is equal to some constant for each i value because x and $x_i$ are*

*both known constants. Substitute $C_i$,*

*where $C_i$ is the product of all the x distances divided by the the x distance $(x - x_i)$*

*Of the individual term resulting in a weighted average of $1/ (x - x_i)$ ;*

$$min(\sum_{i=1}^{n} (Z_\varrho - Y_i)^2 C_i)$$

*Which can be simplified into an equation(same thing as weighted average);*

$$(\sum_{i=1}^{n} (Z_\varrho - Y_i)^2 C_i)$$

$$(\sum_{i=1}^{n} (Z_\varrho - Y_i)^2 C_i)$$

$$\sum_{i=1}^{n} Z_\varrho^2 C_i - \sum_{i=1}^{n} 2Z_\varrho Y_i C_i + \sum_{i=1}^{n} Y_i^2 C_i$$

$$Z_Q{}^2 \sum_{i=1}^{n} C_i - 2Z_Q \sum_{i=1}^{n} Y_i C_i + \sum_{i=1}^{n} Y_i{}^2 C_i$$

$$\frac{dt}{dx}(Z_Q{}^2 \sum_{i=1}^{n} C_i - Z_Q \sum_{i=1}^{n} 2Y_i C_i + \sum_{i=1}^{n} Y_i{}^2 C_i) = 2Z_Q \sum_{i=1}^{n} C_i - 2\sum_{i=1}^{n} Y_i C_i$$

$$2Z_Q \sum_{i=1}^{n} C_i - 2\sum_{i=1}^{n} Y_i C_i = 0$$

$$Z_Q \sum_{i=1}^{n} C_i = \sum_{i=1}^{n} Y_i C_i$$

$$Z_Q = \frac{\sum_{i=1}^{n} Y_i C_i}{\sum_{i=1}^{n} C_i}$$

*Substitute back in $c_i$*

$$Z_Q = \frac{\sum_{i=1}^{n} Y_i(\prod_{i=1}^{n}[(x-x_i)^2])/(x-x_i)^2}{\sum_{i=1}^{n} (\prod_{i=1}^{n}[(x-x_i)^2])/(x-x_i)^2} = \frac{\sum_{i=1}^{n} Y_i(1/(x-x_i)^2)}{\sum_{i=1}^{n} (1/(x-x_i)^2)}$$

This can also can be generalized to any number of dimensions where;

$$min(\sum_{i=1}^{n}(Z_Q - Y_i)^2 / \prod_{a=1}^{b}(D_i - D_{i,a})^2)$$

$$min(\sum_{i=1}^{n}(Z_Q - Y_i)^2 / ((\prod_{i=1}^{n}(\prod_{a=1}^{b}(D_i - D_{i,a})^2))) / \prod_{a=1}^{b}(D_i - D_{i,a})^2)$$

$$min(\sum_{i=1}^{n}(Z_Q - Y_i)^2 k_i)$$

$$Z_Q = \frac{\sum\limits_{i=1}^{n} Y_i \, k_i}{\sum\limits_{i=1}^{n} k_{ii}}$$

$$Z_Q = \frac{\sum\limits_{i=1}^{n} Y_i((\prod\limits_{i=1}^{n}(\prod\limits_{a=1}^{b}(D_i - D_{i,a})^2)))/\prod\limits_{a=1}^{b}(D_i - D_{i,a})^2)}{\sum\limits_{i=1}^{n}((\prod\limits_{i=1}^{n}(\prod\limits_{a=1}^{b}(D_i - D_{i,a})^2)))/\prod\limits_{a=1}^{b}(D_i - D_{i,a})^2)} = Z_Q = \frac{\sum\limits_{i=1}^{n} Y_i(1/\prod\limits_{a=1}^{b}(D_i - D_{i,a})^2)}{\sum\limits_{i=1}^{n}(1/\prod\limits_{a=1}^{b}(D_i - D_{i,a})^2)}$$

This function is both easy to solve and provides an effective weighted function relative to x distance. However, division weights can be ineffective when more white noise is involved because they give extremely high weights to the data closser in x value. This weighted approach is prone to overfitting and can become extreme in semi random functions as seen. In the following graph the division weight model was given ten data points generated through the function and then multiplied by a number 0.5-1.5 with an equal probability distribution. Given these inputs, the model was unable to create a consistent estimate because of its overfitting problem.(the blue points: estimates, orange point: given data)

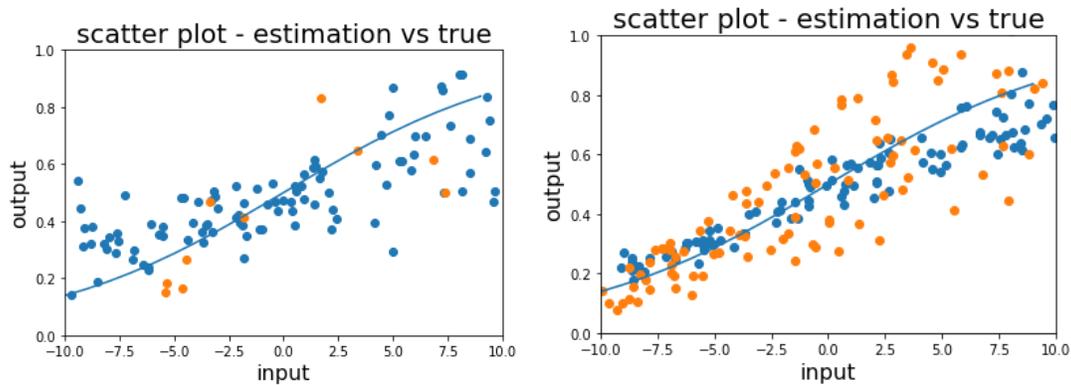

Since in this closed solution of $1/(x - x_i)$ is a constant, any function can replace $1/(x - x_i)$ f$(x - x_i)$ as long as it is in this form. Furthermore, if a constant k is added to the bottom you reduce the variance and overfitting problem. As seen here you can find a closed form solution to any of these functions with any number of variables or dimensions.

$$min(\sum\limits_{i=1}^{n}(Z_Q - Y_i)^2/\prod\limits_{a=1}^{b} f_i((D_i - D_{i,a})^2))$$

$$min(\sum_{i=1}^{n}(Z_Q - Y_i)^2 k_i)$$

**2.3 Optimizing Cost Function** For a weighted distance function of $1/r^{(x-x_i)^2}$ where given the decrease in the value of r, the extremities of the weights for the distance based weights are decreased. When r is decreased, the set of points produced by estimating many values of x become much closer to the average, because when r is increased the weight function grows more quickly and, therefore, the extremities of the weights change as well. Below you can see how r value changes the variance of the predictions:

r =1.2

r =1.8

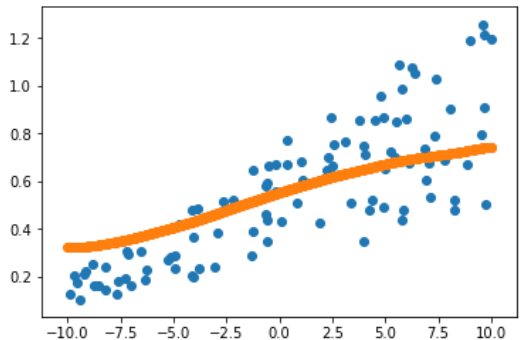
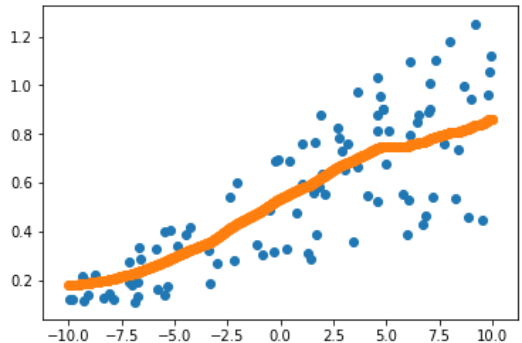

r= 2.6

r =2000

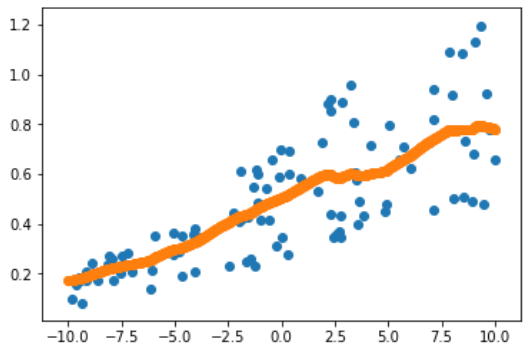
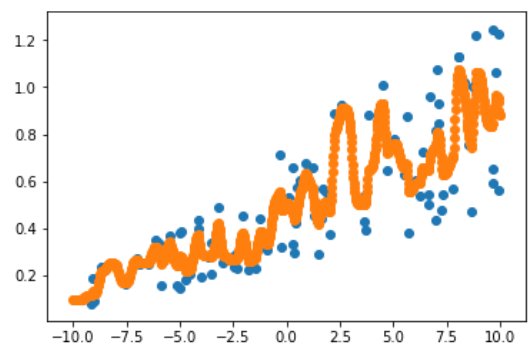

The value of r allows you to decrease or increase the level of overfitting and smoothing. By using this weighting formula instead of using the $1/r^{(x-x_i)^2}$ ; you eliminate undefined points and reduce overfitting and allowing for greater control of the level of variance of the best fit line when compared to the other cost functions. Furthermore, this approach can be optimized to best fit the specific dataset.  When estimating a numerical value in statistics, the goal of the function is to create a predicted best fit curve that has limited variance as well reducing the distance of your estimates to the set of given points. However, if you have a theoretical function that is very complex, it will have high variance. For example, consider a sine curve where your predicted curve needs to match this function. Therefore, in order to match a given dataset's variance, we can adjust our cost function to $1/r^{(x-x_i)^2}$ by adjusting the r value to match the dataset's variance. Using Euler's method, you can iterate to minimize the square of the difference of the variance of the predicted function and the actual function, allowing you to create a more accurate estimation of a best fit line with similar variance by minimizing this function:

$$\left(\frac{\left(\left(\left(\sum_{ii}^{n} Y_i/n\right) - Y_i\right)/n\right)^2}{\left(\left(\left(\sum_{i}^{n} e_i/n\right) - e_i\right)/n\right)^2} - 1\right)^2$$

Where $Y_i$ represents the y values of the given data set and $e_i$ represent the estimates of the X-axis based regression.  This cost function essentially is the average of each set minus the individual points divided by the number of points.  This results in a good approximation of the dataset.

 However, this approach assumes 0 randomness because randomness will increase the variance causing the estimated variance to be artificially high. You can get around this limitation by assuming an initial randomness then estimating and iterating your guesses based on the function:

$$w(r) = \frac{\left(\sum_{i=1}^{n} |(Z_i/Y_i - 1)|\right)}{n}.$$

 However, this approach tends to jump back and forth between points to greater extremes as function of how far away from the actual randomness, so by minimizing the change in predicted randomness you can create a accurate estimation of randomness. In this case I have simply used $1/r^{(x-x_i)^2}$. However, this assumes that 100% of the variance in the data is explained, therefore you can add a parameter to r of what percent of the variance should be explained. Furthermore, this can be expanded to have a greater number of adjustable variables. For example, $1/(r^{(x-x_i)^2} + log_q(x-x)$. However, with more than one variable you have to reduce both the difference between the variance of predicted and actual points as well as the r squared. In the last example there are many values or r and q that achieve the same variance, so we can find value that does not just have matching variance but also has the least r squared. This optimum can be solved by using gradient descent on a weighted cost function where the variance is weighted more than the r squared. This method is more computationally taxing but far superior. This approach weights r squared unlike the one variable and can minimize over more than one variable allowing it to be even more exact.

Given the randomness this model performs quite well when estimating a semi random function as seen in the following graph in which the orange are the given, the blue the estimate, and green the true. However, it should be noted that in my simulation the X-axis based regression had significantly less iterations for Euler's method and used a very basic distance weighting function.

1.
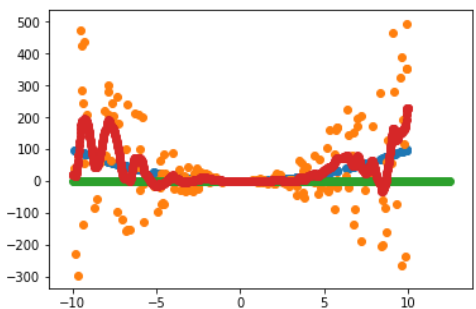

2.
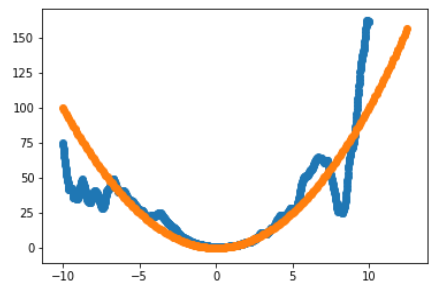

3.
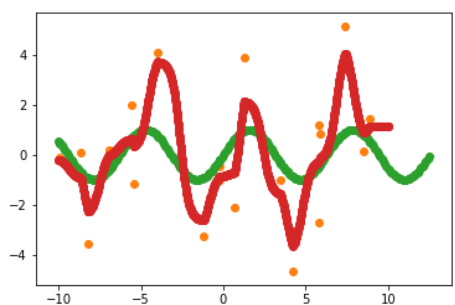

4.
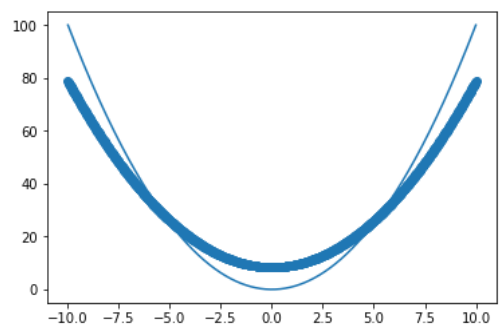

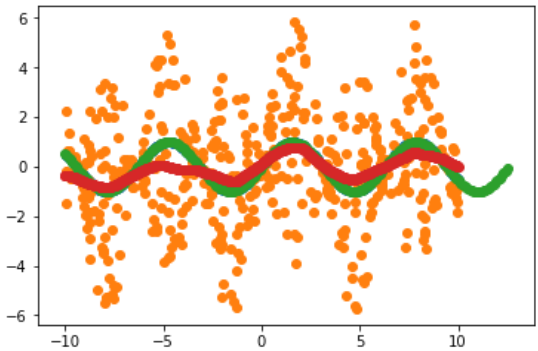

5.

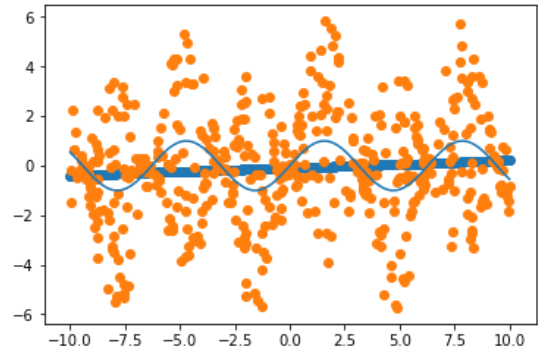

6.

1:X squared iterate randomness   2: Given randomness x squared

3: iterated randomness sin(x)    4: lasso regression x squared

5: sine function X axis regression given randomness  6: polynomial lasso sine

As seen above the polynomial lasso was only effective at estimating the x squared graph while failing completely on the sine wave graph. This is probably because the x squared is a polynomial of degree two, so it would clearly favor polynomial lasso, while the sine which shows the versatility of the x-axis based regression. Polynomial lasso is unlikely to capture functions that do not appear to be like simple polynomials. So in every fair test the X-axis model did much better. X-axis based regression performed well and was significantly outperforming its alternative with an artificial test. X-axis based regression could easily be adapted to more dimensions just by summing the distance weighting variables. Furthermore, x-axis based regression only became ineffective at the edges where there is significantly less data.

**3.1 Testing on "Real" Data Sets** This specific example will illustrate the effectiveness of this new tool. The system can predict outcomes in the real world. This new forecasting technique is highly flexible and can be effectively applied to a vast array of other data sets from biological systems, to economics, to astrophysics and many others. It can be applied when predicting almost any system in which we currently use regression or machine learning to predict numerical classifiers.

I will first employ traditional regressions to estimate crop yield in Africa by considering both last previous year's yields and the amount of land harvested. I will then compare results of this legacy method to the results I generate from my new approach in order to find patterns in the data and to find future outlooks on crop yields.

I decided to look at agricultural production in Africa because of its extremely low agricultural yields, and was wondering if there was a way that we could address this problem before this imminent threat becomes a crisis in sub Saharan Africa.

In the OECD-FAO Agricultural Outlook 2018-2027, the two UN organizations highlight basic problems of sub-Saharan agriculture, "Despite accounting for over 13% of the world population and close to 20% of global agricultural land, sub-Saharan Africa's share of global agricultural output is relatively low," the OECD-FAO said. "Agricultural production is constrained by challenging agro-ecological conditions, limited access to and utilization of technology, and the fact that economic growth in many cases remains only marginally ahead of population increases."

In the graphs below you can see the estimated best fit lines for 1000 data points when comparing crop yields 1 year ago, the X-axis, to how much they increased over this 1 year period (normalized), the Y-axis, for cereal products, the most significant agricultural staple in Africa. The graph on the right is the X-axis model and the one on the left is the lasso regression.

Polynomial lasso regression                      X-axis based regression

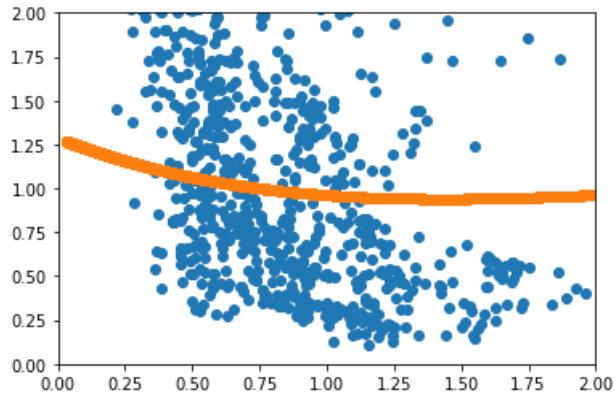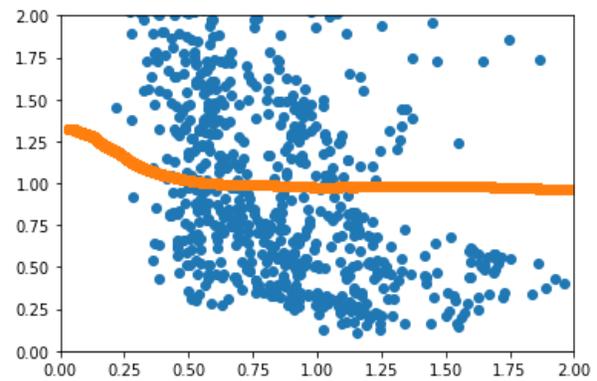

The accuracy of the X-axis regression is 4.6% better, with 1000 points. The X-axis regression beats out the traditional polynomial regression because the X-axis best fit line is not limited to a polynomial curve and can capture the the minute intricacy of the relation. This can be seen with the slight ups and downs from 0.5 to 2.0 and in the beginning where the X-axis is able to predict a slightly more complex curve. This further validates the enhanced effectiveness of the X-axis weighted regression when applied to real world problems. By using a more accurate model to predict future crop yields, countries can better allocate resources, manage reserves, and alert foreign governments about potential aid needs. Furthermore, these results support the models effectiveness in predicting crop yields but also its applicability to real world problems.

| Number of points | Total iterations | Lasso error | X-axis error | Percent advantage |
|---|---|---|---|---|
| 1000 | 5000 | 0.16057744 | 0.15342135612367921 | 4.66433% |
| 2000 | 5000 | 0.15614923 | 0.1474352255852411 | 5.91039% |
| 2750 | 5000 | 0.15644323 | 0.1497225507429759 | 4.48875% |

# Conclusion:

In conclusion, the X-axis regression is more accurate both in artificial and real world tests than the lasso regression, which is widely considered to be one of the best regressions. The X-axis based regression delivers much better predictions and can be applied more widely than tested in the paper; some examples include, fiscal policy, medicine effectiveness, and physics. I would love to see future work in the following areas 1.) determining the amount of explained variance 2.) a better function to quantify randomness and 3.) more application of this regression to real world problems.